\documentclass[pdflatex,sn-mathphys-num]{sn-jnl}% Math and Physical Sciences Numbered Reference Style
\usepackage{graphicx}%
\usepackage{multirow}%
\usepackage{amsmath,amssymb,amsfonts}%
\usepackage{amsthm}%
\usepackage{mathrsfs}%
\usepackage[title]{appendix}%
\usepackage{xcolor}%
\usepackage{textcomp}%
\usepackage{manyfoot}%
\usepackage{booktabs}%
\usepackage{algorithm}%
\usepackage{algorithmicx}%
\usepackage{algpseudocode}%
\usepackage{listings}%
\usepackage[caption=false,font=footnotesize]{subfig}
\usepackage{caption}
\usepackage{adjustbox}
\usepackage{booktabs}

\theoremstyle{thmstyleone}%
\theoremstyle{thmstyletwo}%

\theoremstyle{thmstylethree}%

\begin{document}

\title[Article Title]{Enhancing Diagnostic Accuracy for Urinary Tract Disease through Explainable SHAP-Guided Feature Selection and Classification}
% {Seleção de características com valores SHAP para auxílio ao diagnóstico de doenças do trato urinário}


\author[1]{\fnm{Filipe Ferreira} \spfx{de} \sur{Oliveira}}\email{filipe.f.oliveira98@gmail.com}

\author[1,2]{\fnm{Matheus Becali} \sur{Rocha}}\email{matheusbecali@gmail.com}

\author*[1,2]{\fnm{Renato A.} \sur{Krohling}}\email{krohling.renato@gmail.com}

\affil[1]{\orgdiv{Labcin - Nature Inspired Computing Laboratory}, \orgname{Federal University of Esp\'irito Santo}, \orgaddress{\city{Vit\'oria}, \country{Brazil}}}

\affil[2]{\orgdiv{Graduate Program in Computer Science}, \orgname{Federal University of Esp\'irito Santo}, \orgaddress{\city{Vit\'oria}, \country{Brazil}}}


% \abstract{The abstract serves both as a general introduction to the topic and as a brief, non-technical summary of the main results and their implications. Authors are advised to check the author instructions for the journal they are submitting to for word limits and if structural elements like subheadings, citations, or equations are permitted.}

%%================================%%
%% Sample for structured abstract %%
%%================================%%

\abstract{
\textbf{Purpose:} To propose an approach to support the diagnosis of urinary tract diseases, with a focus on bladder cancer, using SHAP (SHapley Additive exPlanations)-based feature selection to enhance the transparency and effectiveness of predictive models.

\textbf{Methods:} Six binary classification scenarios were developed to distinguish bladder cancer from other urological and oncological conditions. The algorithms XGBoost, LightGBM, and CatBoost were employed, with hyperparameter optimization performed using Optuna and class balancing with the SMOTE technique. The selection of predictive variables was guided by importance values calculated through SHAP.

\textbf{Results:} SHAP-based feature selection efficiently reduced data dimensionality while maintaining or even improving performance metrics such as balanced accuracy, precision, and specificity.

\textbf{Conclusion:} The use of explainability techniques (SHAP) for feature selection proved to be an effective approach. The proposed methodology may contribute to the development of more transparent, reliable, and efficient clinical decision support systems, optimizing screening and early diagnosis of urinary tract diseases.
}

% \abstract{\textbf{Purpose:} Propor uma abordagem de auxílio ao diagnóstico de doenças do trato urinário, com foco em câncer de bexiga, utilizando seleção de características baseada em SHAP (SHapley Additive exPlanations), visando aumentar a transparência e a eficácia dos modelos preditivos.

% \textbf{Methods:} Foram desenvolvidos seis cenários de classificação binária para distinguir câncer de bexiga de outras condições urológicas e oncológicas. Foram empregados os algoritmos XGBoost, LightGBM e CatBoost, com otimização de hiperparâmetros realizada pelo Optuna e balanceamento de classes pela técnica SMOTE. A seleção de variáveis preditivas foi guiada pelos valores de importância calculados via SHAP.

% \textbf{Results:} A seleção de características via SHAP reduziu eficientemente a dimensionalidade dos dados, mantendo ou até melhorando as métricas de desempenho, como acurácia balanceada, precisão e especificidade.

% \textbf{Conclusion:} A utilização de técnicas de explicabilidade (SHAP) para seleção de variáveis apresentou-se como uma abordagem eficiente. A metodologia proposta pode contribuir para o desenvolvimento de sistemas de apoio à decisão clínica mais transparentes, confiáveis e eficientes, otimizando a triagem e o diagnóstico precoce de doenças do trato urinário.
% }


% MAX 250 WORDS



\maketitle

\section{Introduction}\label{sec1}

% English
Artificial Intelligence (AI) has rapidly expanded across various domains, including the medical field \cite{Davenport2018}. Recent examples include the use of machine learning algorithms for the diagnosis of COVID-19 \cite{Krohling2023} and the detection of skin lesions \cite{Vidya2020}, highlighting the potential of these technologies in automating decision-making processes. However, many of these models, particularly those based on deep learning, operate as black boxes, providing high accuracy but low transparency regarding the contribution of input variables to the final outcome \cite{Rudin2019}. This lack of interpretability is especially critical in high-risk contexts such as clinical diagnosis, with bladder cancer (BCa) being a prominent example. Classified as the tenth most common type of neoplasm and the thirteenth leading cause of cancer-related death worldwide \cite{Zhang2023}, with more than 90\% of cases belonging to the urothelial carcinoma subtype \cite{vanHoogstraten2023}, BCa presents both high incidence and heterogeneous clinical behavior. In this scenario, improving early diagnostic methods is essential, making the application of transparent and interpretable AI models crucial to reducing the morbidity and mortality associated with this neoplasm.

% Portuguese
% A Inteligência Artificial (IA) tem se expandido rapidamente em diversos domínios, incluindo a área médica \cite{Davenport2018}. Exemplos recentes incluem a utilização de algoritmos de aprendizado de máquina no diagnóstico de COVID-19 \cite{Krohling2023} e na detecção de lesões cutâneas \cite{Vidya2020}, evidenciando o potencial dessas tecnologias na automação da tomada de decisão. Contudo, muitos desses modelos, em especial os de aprendizado profundo, funcionam como \textit{caixas-pretas}, apresentando elevada acurácia, mas baixa transparência quanto à contribuição das variáveis de entrada para o resultado final \cite{Rudin2019}. Essa falta de interpretabilidade é especialmente crítica em contextos de alto risco, como o diagnóstico clínico, e um exemplo proeminente desse desafio é o câncer de bexiga (BCa). Classificado como o décimo tipo de neoplasia mais frequente e a décima terceira principal causa de morte por câncer em âmbito global \cite{Zhang2023}, com mais de 90\% dos casos sendo do subtipo carcinoma urotelial \cite{vanHoogstraten2023}, o BCa possui elevada incidência e comportamento clínico heterogêneo. Nesse cenário, o aprimoramento dos métodos de diagnóstico precoce é fundamental, tornando a aplicação de modelos de IA transparentes e interpretáveis essencial para reduzir a morbidade e a mortalidade associadas a essa neoplasia.

% #####################################

% English
Smoking is identified as the main risk factor, being associated with approximately half of the cases and 37\% of disease-related deaths \cite{Zhang2023}. Occupational exposure to carcinogenic substances represents the second most relevant factor, accounting for about 10\% of cases \cite{Zhang2023}, which contributes to an incidence up to four times higher in men compared to women. Lifestyle and environment-related factors also play a role, as the urothelial epithelium is continuously exposed to potentially mutagenic substances present in the urine \cite{HalasehSattamA2022}. The most frequently reported symptom is hematuria (blood in the urine), which may be macroscopic (visible) or microscopic (detected through laboratory tests), typically intermittent and painless \cite{Pricee584, dulku2019_hematuria}.

% Portuguese
% O tabagismo é identificado como o principal fator de risco, estando associado a aproximadamente metade dos casos e a 37\% dos óbitos relacionados à doença \cite{Zhang2023}. A exposição ocupacional a substâncias carcinogênicas constitui o segundo fator mais relevante, sendo responsável por cerca de 10\% dos casos \cite{Zhang2023}, o que contribui para a incidência até quatro vezes maior em homens quando comparada à das mulheres. Fatores relacionados ao estilo de vida e ao ambiente também exercem influência, uma vez que o epitélio do trato urinário está continuamente exposto a substâncias potencialmente mutagênicas presentes na urina \cite{HalasehSattamA2022}. O sintoma mais frequentemente relatado é a hematúria (presença de sangue na urina), que pode ser macroscópica (visível) ou microscópica (detectada por exames laboratoriais), geralmente de caráter intermitente e não acompanhada de dor \cite{Pricee584, dulku2019_hematuria}.

% English
Beyond its clinical impact, bladder cancer imposes a significant financial burden on healthcare systems due to its high recurrence rate and the costs of diagnostic procedures, such as periodic cystoscopies, intravesical treatments, and more complex surgeries \cite{vanHoogstraten2023, HalasehSattamA2022}. In developed regions, where population aging is increasing, the global burden of BCa is projected to rise in the coming decades, underscoring the need for improved strategies for prevention, diagnosis, and treatment \cite{Zhang2023}.

% Portuguese
% Além do impacto clínico, o câncer de bexiga acarreta elevado custo aos sistemas de saúde devido à alta taxa de recidiva e ao valor dos procedimentos diagnósticos, como cistoscopias periódicas, tratamentos intravesicais e cirurgias de maior complexidade \cite{vanHoogstraten2023, HalasehSattamA2022}. Em regiões desenvolvidas, onde o envelhecimento populacional é crescente, projeta-se que a carga global de BCa aumente nas próximas décadas, reforçando a necessidade de estratégias aprimoradas de prevenção, diagnóstico e tratamento \cite{Zhang2023}.

% English
In this context, the development of more accurate and interpretable models and analytical methods may play a relevant role in early detection and therapeutic management. Approaches that enhance the understanding of the determinants of bladder cancer progression have the potential to optimize clinical decision-making and reduce the associated morbidity and mortality.

% Portuguese
% Nesse contexto, o desenvolvimento de modelos e métodos de análise mais precisos e interpretáveis pode desempenhar papel relevante na detecção precoce e no manejo terapêutico. Abordagens que favoreçam a compreensão dos fatores determinantes na progressão do câncer de bexiga têm potencial para otimizar a tomada de decisão clínica e reduzir a morbimortalidade associada.
% ######################################

% English
Clinical and epidemiological databases often present high dimensionality, with many variables and, at the same time, a relatively small number of samples \cite{Hastie2009}. Under these conditions, conventional machine learning algorithms may produce models with limited generalization power. Dimensionality reduction techniques, such as Principal Component Analysis (PCA) \cite{Pearson1901} and SHAP (SHapley Additive exPlanations) \cite{Lundberg2017}, help mitigate this issue but do not fully address the challenge of interpretability \cite{Rudin2019}.

% Portuguese
% As bases de dados clínicos e epidemiológicos frequentemente apresentam alta dimensionalidade, muitas variáveis e, simultaneamente, número reduzido de amostras \cite{Hastie2009}. Nessas condições, algoritmos de aprendizado de máquina convencionais podem gerar modelos com baixo poder de generalização. Técnicas de redução de dimensionalidade, como a Análise de Componentes Principais (PCA, do inglês \textit{Principal Component Analysis}) \cite{Pearson1901} e SHAP (\textit{SHapley Additive exPlanations}) \cite{Lundberg2017}, contribuem para mitigar o problema, mas não solucionam totalmente a questão da interpretabilidade \cite{Rudin2019}.

% English
This study proposes the application of SHAP for dimensionality reduction. The aim is to identify potential relationships between variables with high predictive importance and the outcome of interest. For the analysis, a dataset consisting of 1,336 clinical and laboratory samples from patients with different urinary tract diseases, collected by \cite{Tsai2022} between 2017 and 2022, will be used.

% Portuguese
% Este trabalho propõe a aplicação do SHAP para redução de dimensionalidade. O objetivo é identificar possíveis relações entre variáveis de elevada importância preditiva e o desfecho de interesse. Para o estudo, será utilizada uma base de dados composta por 1336 amostras clínico-laboratoriais de pacientes com diferentes doenças do trato urinário, coletada por \cite{Tsai2022} entre 2017 e 2022.

% English
Although SHAP was originally developed for feature interpretation, it can also be applied to dimensionality reduction. Kumar et al. (2020) \cite{Kumar2020DRShap} employed this technique on two datasets, demonstrating that, in addition to assigning importance to variables, it enables the simplification of complex models without compromising the reliability of predictions.

% Portuguese
% Embora o SHAP tenha sido originalmente desenvolvido para interpretação de características, também pode ser empregado para redução de dimensionalidade. Kumar et al. (2020) \cite{Kumar2020DRShap} aplicaram essa técnica em duas bases de dados, mostrando que, além de atribuir importância às variáveis, ela permite simplificar modelos complexos sem comprometer a confiabilidade das previsões.

% English
Liu et al. (2022) \cite{LIU2022shap} employed machine learning models combined with SHAP for feature selection in the diagnosis of Parkinson’s disease, integrating algorithms such as Deep Forest (gcForest), Extreme Gradient Boosting (XGBoost), Light Gradient Boosting Machine (LightGBM), and Random Forest (RF). The results indicated superior performance compared to conventional techniques, achieving accuracy above 91\% and an F1-score of 0.945.

% Portuguese
% Liu et al. (2022) \cite{LIU2022shap} utilizaram modelos de aprendizado de máquina aliados ao SHAP na seleção de características para diagnóstico da doença de Parkinson, combinando algoritmos como \textit{Deep Forest} (gcForest), \textit{Extreme Gradient Boosting} (XGBoost), \textit{Light Gradient Boosting Machine} (LightGBM) e \textit{Random Forest} (RF). Os resultados indicaram desempenho superior ao de técnicas convencionais, alcançando acurácia acima de 91\% e \textit{F1-score} de 0,945.

% English
The structure of this article is organized as follows: Section~\ref{sec:MatMethods} presents in detail the data used in the study, the machine learning models employed, and the experimental methodology adopted; Section~\ref{sec:results} describes the experimental setup, the results obtained, and the associated discussion. Subsequently, Section~\ref{sec:Discussion} addresses the critical analysis of the results and relevant interpretations. Finally, Section~\ref{sec:conclusion} provides the study’s conclusions and highlights possible directions for future work.

% Portuguese
% A estrutura do artigo segue o seguinte formato: na Seção~\ref{sec:MatMethods} apresenta detalhadamente os dados utilizados no estudo, os modelos de aprendizado de máquina empregados e a metodologia experimental adotada; na Seção~\ref{sec:results} descreve a configuração experimental, os resultados obtidos e a discussão associada. Em seguida, a Seção~\ref{sec:Discussion} aborda a análise crítica dos resultados e interpretações relevantes. Por fim, a Seção~\ref{sec:conclusion} apresenta as conclusões do estudo e aponta possíveis direções para trabalhos futuros.

\section{Material and Methods}\label{sec:MatMethods}

\subsection{Patient data}\label{descpro}

% English
The dataset used consists of clinical and laboratory records, including urinalysis and biochemical tests, collected at Mackay Memorial Hospital between January 2017 and February 2020 \cite{Tsai2022}. It comprises 1,336 samples distributed across five classes: 591 cases of bladder cancer, 201 cases of prostate cancer, 200 cases of kidney cancer, 200 cases of uterine cancer, and 144 cases of cystitis, as shown in Table~\ref{tbl-classes-dataset}. The dataset contains a total of 39 variables, a considerable number that reinforces the need for dimensionality reduction. % The variables included in the dataset are listed in Table~\ref{tbl-features}, indicating their type and the percentage of missing values. The data were collected and made available by \cite{Tsai2022}.

% Portuguese
% O conjunto de dados utilizado consiste em registros clínico-laboratoriais, incluindo exames de urina e testes bioquímicos, coletados no Mackay Memorial Hospital entre janeiro de 2017 e fevereiro de 2020 \cite{Tsai2022}. A base contém 1336 amostras distribuídas em cinco classes: 591 casos de câncer de bexiga, 201 casos de câncer de próstata, 200 casos de câncer de rins, 200 casos de câncer de útero e 144 casos de cistite, conforme apresentado na Tabela~\ref{tbl-classes-dataset}. O conjunto contém um total 39 variáveis, um valor significativo que reforça a necessidade da redução de dimensionalidade. % As variáveis presentes no conjunto de dados estão listadas na Tabela~\ref{tbl-features} indicando o seu tipo e a porcentagem de dados faltantes. Os dados foram coletados e disponibilizados por \cite{Tsai2022}. 

\begin{table}[htbp]
\caption{Class distribution of the dataset.}
\label{tbl-classes-dataset}
\centering
\begin{tabular}{ c  c  c }\hline
\textbf{Disease} & \textbf{Number of samples} & \textbf{Missing data (\%)}\\\hline
Bladder cancer  & 591 & 42,27 \\ 
Prostate cancer & 201 & 15,04 \\ 
Kidney cancer   & 200 & 14,97 \\ 
Uterus cancer   & 200 & 14,97 \\ 
Cystitis        & 144 & 10,78 \\ \hline
\end{tabular}
\end{table}

\subsection{Data Preprocessing}
\subsubsection{Missing Data Imputation}
\label{sec-imputacao}

% English
The treatment of missing values is a critical step in data preprocessing, as failures in extraction or collection may generate gaps that compromise model performance \cite{Aljrees2024knn}. Proper imputation helps reduce biases and preserve the consistency of subsequent analyses. In this study, the KNNImputer, based on the K-Nearest Neighbors (KNN) algorithm, was employed \cite{batista2002knnimp}. The method estimates missing values from the nearest samples in the feature space, assuming that similar instances exhibit correlated patterns \cite{Aljrees2024knn}. The procedure follows three main steps: (i) for each instance with missing values, a set of $k$ nearest neighbors is identified according to a distance metric, such as Euclidean distance; (ii) the known values of these neighbors are used to estimate the missing value; (iii) the imputation is carried out using the mean, median, or mode, depending on the variable type and the adopted configuration.

% Portuguese
% O tratamento de valores ausentes constitui uma etapa crítica do pré-processamento de dados, uma vez que falhas na extração ou coleta podem gerar lacunas que comprometem o desempenho dos modelos \cite{Aljrees2024knn}. A imputação adequada contribui para reduzir vieses e preservar a consistência das análises subsequentes. Neste trabalho, foi utilizado o \textit{KNNImputer}, baseado no algoritmo \textit{K-Nearest Neighbors} (KNN) \cite{batista2002knnimp}. O método estima valores ausentes a partir das amostras mais próximas no espaço das variáveis, considerando que instâncias semelhantes apresentam padrões correlacionados \cite{Aljrees2024knn}. O procedimento segue três etapas principais: (i) para cada instância com valores ausentes, identifica-se um conjunto de \(k\) vizinhos mais próximos, segundo uma métrica de distância, como a euclidiana; (ii) os valores conhecidos desses vizinhos são utilizados para estimar o valor ausente; (iii) a imputação é realizada por meio da média, mediana ou moda, dependendo do tipo de variável e da configuração adotada.

\subsubsection{Oversampling}
\label{sec-smote}

% English
Imbalanced datasets, in which some classes are underrepresented, can hinder the generalization of machine learning algorithms for those classes \cite{He2009LearningFI}. To address this issue, the Synthetic Minority Over-sampling Technique (SMOTE) \cite{Chawla_2002_Smote} was applied, which generates new synthetic samples instead of simply replicating minority instances. The method selects a minority class instance, identifies its $k$ nearest neighbors, randomly chooses one of them, and creates a new sample at a random point along the line segment connecting the original instance to the neighbor; this process is repeated until the desired class proportion is achieved \cite{fernandez2018learning}.

% Portuguese
% Bases de dados desbalanceadas, nas quais algumas classes estão sub-representadas, podem prejudicar a generalização de algoritmos de aprendizado de máquina para essas classes \cite{He2009LearningFI}. Para contornar esse problema, utilizou-se a técnica Synthetic Minority Over-sampling Technique (SMOTE) \cite{Chawla_2002_Smote}, que gera novas amostras sintéticas em vez de simplesmente replicar instâncias minoritárias. O método seleciona uma instância da classe minoritária, identifica seus $k$ vizinhos mais próximos, escolhe um deles aleatoriamente e cria uma nova amostra em um ponto aleatório ao longo do segmento que conecta a instância original ao vizinho; esse processo se repete até alcançar a proporção desejada de classes \cite{fernandez2018learning}.

\subsubsection{Dimensionality Reduction}
\label{sec-reduc-dimens}

% English
In high-dimensional datasets, the presence of irrelevant or redundant variables can degrade the performance of machine learning algorithms, increase computational cost, and reduce model interpretability \cite{Wilson2020, LIU2022shap}. To mitigate these effects, feature selection methods such as SHapley Additive exPlanations (SHAP) \cite{Lundberg2017} allow the identification and prioritization of variables with greater predictive impact. SHAP decomposes each prediction into the sum of the individual contributions of the variables, providing explainability and supporting feature selection. Although the exact calculation based on Shapley values is infeasible in high-dimensional problems, SHAP employs computational approximations, such as weighted linear regression for generic models and specific optimizations for tree-based models \cite{Wilson2021DR}. The SHAP value of a variable $i$ indicates its average marginal contribution to the prediction, considering all possible orders of inclusion of the variables in the model.

% Portuguese
% Em conjuntos de dados de alta dimensionalidade, a presença de variáveis irrelevantes ou redundantes pode degradar o desempenho de algoritmos de aprendizado de máquina, aumentar o custo computacional e reduzir a interpretabilidade dos modelos \cite{Wilson2020, LIU2022shap}. Para mitigar esses efeitos, métodos de seleção de características, como o SHapley Additive exPlanations (SHAP) \cite{Lundberg2017}, permitem identificar e priorizar variáveis com maior impacto preditivo. O SHAP decompõe cada previsão na soma das contribuições individuais das variáveis, oferecendo explicabilidade e suporte à seleção de características. Embora o cálculo exato baseado nos valores de Shapley seja inviável em problemas de alta dimensão, o SHAP utiliza aproximações computacionais, como regressão linear ponderada para modelos genéricos e otimizações específicas para modelos baseados em árvores \cite{Wilson2021DR}. O valor SHAP de uma variável $i$ indica sua contribuição marginal média para a previsão, considerando todas as ordens possíveis de inclusão das variáveis no modelo.

\subsection{Algorithms}\label{subsec:AlgMethods}

%%=====================================================================================%%

\subsubsection{Extreme Gradient Boosting (XGBoost)}
\label{sec-xgboost}

% English
eXtreme Gradient Boosting (XGBoost) \cite{Chen2016xgboost} is a tree-based Gradient Boosting algorithm, recognized for its high performance, scalability, and efficient regularization for overfitting control. The boosting technique consists of iteratively combining weak models (typically shallow decision trees) to form a robust model.

% Portuguese
% O eXtreme Gradient Boosting (XGBoost) \cite{Chen2016xgboost}, é um algoritmo de \textit{Gradient Boosting} baseado em árvores de decisão, reconhecido por sua alta performance, escalabilidade e regularização eficiente para controle de sobreajuste. A técnica de \textit{boosting} consiste em combinar iterativamente modelos fracos (geralmente árvores de decisão rasas) para formar um modelo robusto. 

%%=====================================================================================%%

\subsubsection{Categorical Boosting (CatBoost)}
\label{sec-catboost}

% English
Categorical Boosting (CatBoost) \cite{Prokhorenkova2017CatBoost} is a Gradient Boosting algorithm based on decision trees that directly handles categorical variables, eliminating the need for manual encoding. It employs Ordered Target-Based Encoding to prevent target leakage and incorporates regularization to reduce overfitting, making it particularly efficient in problems with many categorical variables, such as recommendation systems and demand forecasting.

% Portuguese
% O Categorical Boosting (CatBoost) \cite{Prokhorenkova2017CatBoost} é um algoritmo de \textit{Gradient Boosting} sobre árvores de decisão que lida diretamente com variáveis categóricas, dispensando codificação manual. Utiliza \textit{Ordered Target-Based Encoding} para evitar vazamento de informação do alvo e incorpora regularização para reduzir sobreajuste, sendo especialmente eficiente em problemas com muitas variáveis categóricas, como sistemas de recomendação e previsão de demanda.

%%=====================================================================================%%

\subsubsection{Light Gradient Boosting Machine (LightGBM)}
\label{sec-lightgbm}

% English
The Light Gradient Boosting Machine (LightGBM) \cite{Ke2017LightGBM} is a Gradient Boosting algorithm optimized for large-scale data and high dimensionality. It employs Gradient-Based One-Side Sampling (GOSS) to prioritize instances with larger gradients and reduce sample size without information loss. In addition, it adopts a leaf-wise growth strategy, which expands the leaf that yields the greatest error reduction, requiring regularization to prevent overfitting. These optimizations make training faster and more efficient, making LightGBM particularly suitable for large-scale applications such as fraud detection and recommendation systems.

% Portuguese
% O Light Gradient Boosting Machine (LightGBM) \cite{Ke2017LightGBM} é um algoritmo de \textit{Gradient Boosting} otimizado para grandes volumes de dados e alta dimensionalidade. Emprega \textit{Gradient-Based One-Side Sampling} (GOSS) para priorizar instâncias com maiores gradientes e reduz o tamanho da amostra sem perda de informação, além de adotar expansão \textit{leaf-wise}, que cresce a folha que mais reduz o erro, exigindo regularização para evitar sobreajuste. Essas otimizações tornam o treinamento mais rápido e eficiente, sendo indicado para aplicações em larga escala, como detecção de fraudes e sistemas de recomendação.

%%=====================================================================================%%

\subsection{Experimental Methodology}

% English
The experimental methodology adopted was inspired by the study of Tsai et al. (2022) \cite{Tsai2022}, who used clinical and laboratory data for bladder cancer prediction. Similarly, this work employs a binary classification strategy, defining multiple experimental scenarios, each designed to discriminate between a specific pair of pathological conditions. The scenarios analyzed were:
(1) Bladder Cancer vs. Prostate Cancer, (2) Bladder Cancer vs. Cystitis, (3) Bladder Cancer vs. Kidney Cancer, (4) Bladder Cancer vs. Uterus Cancer, (5) Bladder Cancer vs. Others, and (6) Prostate Cancer vs. Others. In experiments 5 and 6, the goal was to identify cases of bladder or prostate cancer against the other conditions in the dataset, where the class distribution in each scenario is presented in Table~\ref{tbl-exp}.

% Portuguese
% A metodologia experimental adotada foi inspirada no estudo de Tsai et al. (2022) \cite{Tsai2022}, que utilizaram dados clínico-laboratoriais para a predição de câncer de bexiga. De forma análoga, este trabalho emprega uma estratégia de classificação binária, definindo múltiplos cenários experimentais, cada um destinado a discriminar um par específico de condições patológicas. Os cenários analisados foram:  
% (1) Câncer de Bexiga vs. Câncer de Próstata, (2) Câncer de Bexiga vs. Cistite, (3) Câncer de Bexiga vs. Câncer de Rim, (4) Câncer de Bexiga vs. Câncer de Útero, (5) Câncer de Bexiga vs. Outros e (6) Câncer de Próstata vs. Outros. Nos experimentos 5 e 6, o objetivo foi identificar casos de câncer de bexiga ou próstata frente às demais condições do conjunto de dados, onde a distribuição das classes em cada cenário está listada na Tabela~\ref{tbl-exp}.

\begin{table}[htbp]
    \caption{Class Distribution by Experiment}
    \scriptsize
    \label{tbl-exp}
    \centering
\begin{tabular}{ccc}
        \hline
        \textbf{Experiment} & \textbf{Groups}                                                                 & \textbf{Number of Samples} \\ \hline
        1 & \begin{tabular}[c]{@{}c@{}}Bladder Cancer \\ Prostate Cancer\end{tabular} & \begin{tabular}[c]{@{}c@{}}591 \\ 201\end{tabular} \\ \hline
        2 & \begin{tabular}[c]{@{}c@{}}Bladder Cancer \\ Cystitis\end{tabular}       & \begin{tabular}[c]{@{}c@{}}591 \\ 144\end{tabular} \\ \hline
        3 & \begin{tabular}[c]{@{}c@{}}Bladder Cancer \\ Kidney Cancer\end{tabular}  & \begin{tabular}[c]{@{}c@{}}591 \\ 200\end{tabular} \\ \hline
        4 & \begin{tabular}[c]{@{}c@{}}Bladder Cancer \\ Uterus Cancer\end{tabular} & \begin{tabular}[c]{@{}c@{}}591 \\ 200\end{tabular} \\ \hline
        5 & \begin{tabular}[c]{@{}c@{}}Bladder Cancer \\ Others\end{tabular}         & \begin{tabular}[c]{@{}c@{}}591 \\ 745\end{tabular} \\ \hline
        6 & \begin{tabular}[c]{@{}c@{}}Prostate Cancer \\ Others\end{tabular}        & \begin{tabular}[c]{@{}c@{}}201 \\ 1135\end{tabular} \\ \hline
\end{tabular}
\end{table}

\begin{figure}[htbp]
    \centering
    \includegraphics[width=13.1cm]{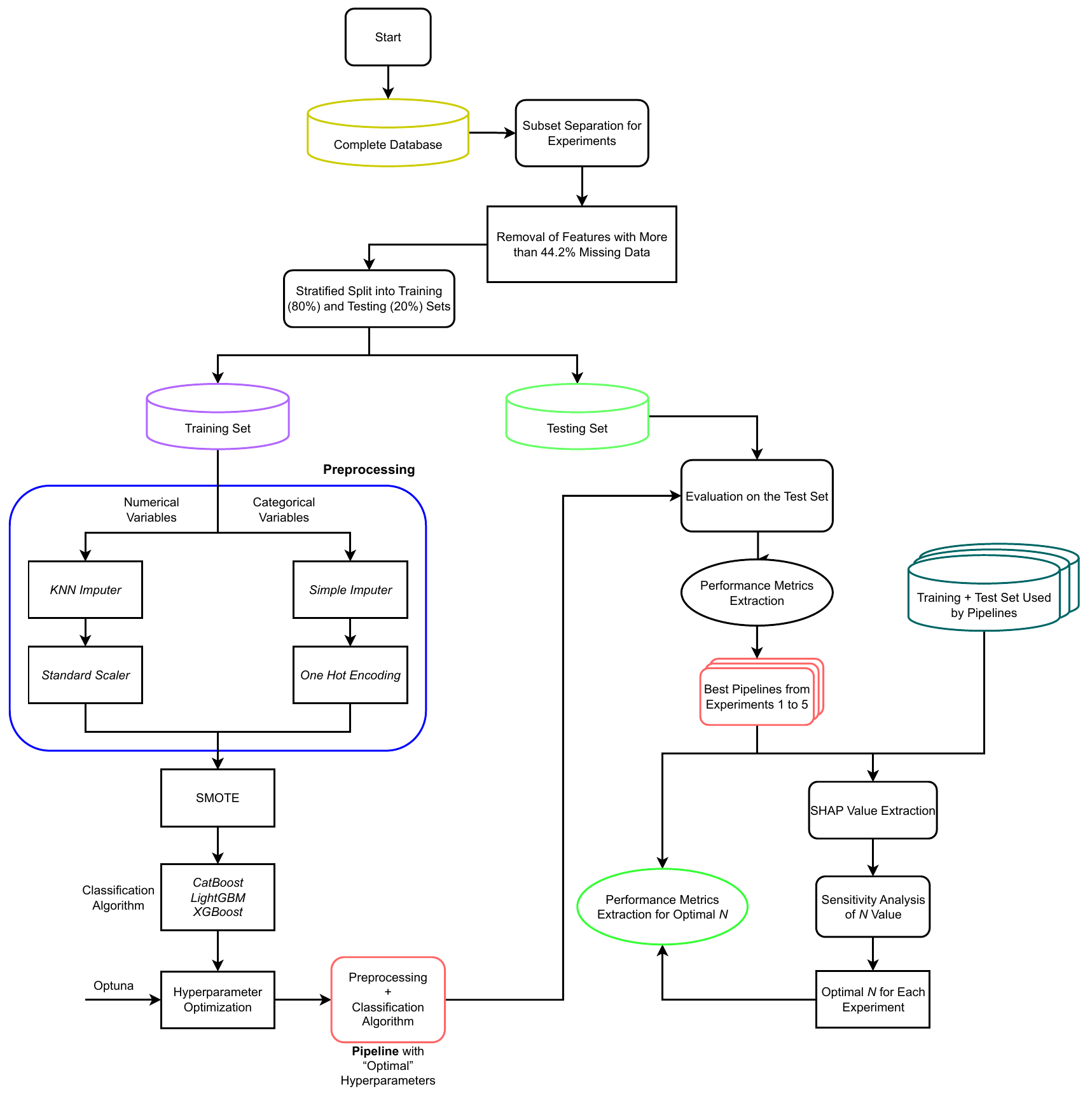}
    \caption{General framework flow: the dataset is divided according to experiments and classes. The training set undergoes preprocessing (imputation, normalization, and encoding) and oversampling with SMOTE, followed by hyperparameter optimization using Optuna. With the optimal parameters, the model is trained and evaluated, generating performance metrics. Subsequently, SHAP values are used for dimensionality reduction and the selection of the most relevant features for classification.}
    \label{fig:pipeline}
\end{figure}

% English
A pipeline was developed to integrate all preprocessing steps and the classification model, serving as the central object of optimization and training. Figure~\ref{fig:pipeline} illustrates its structure. Data preprocessing involved imputing missing values for numerical variables using the KNNImputer, with the number of neighbors $k$ defined through optimization with Optuna, followed by standardization with StandardScaler. For categorical variables, imputation was performed using the mode with SimpleImputer, followed by One-Hot Encoding, resulting in a total of 56 variables for the complete model. Class balancing was applied exclusively to the training data during cross-validation, using the SMOTE technique. The classification step employed machine learning algorithms, including XGBoost, LightGBM, and CatBoost.

% Portuguese
% Foi desenvolvido um \textit{pipeline} para integrar todas as etapas de pré-processamento e o modelo de classificação, sendo o objeto central da otimização e treinamento. A Figura~\ref{fig:pipeline} apresenta sua estrutura. O pré-processamento dos dados envolveu a imputação de valores ausentes para variáveis numéricas utilizando o KNNImputer, com o número de vizinhos $k$ definido por meio de otimização via Optuna, seguido de padronização por StandardScaler. Para as variáveis categóricas, a imputação foi realizada pela moda utilizando SimpleImputer, seguida de codificação por One-Hot Encoding, resultando num total de 56 variáveis para o modelo completo. O balanceamento das classes foi aplicado exclusivamente aos dados de treinamento durante a validação cruzada, utilizando a técnica SMOTE. A etapa de classificação empregou algoritmos de aprendizado de máquina, incluindo XGBoost, LightGBM e CatBoost.

% \begin{itemize}
%     \item \textbf{Pré-processamento}:
%     \begin{itemize}
%         \item Variáveis numéricas tiveram valores ausentes imputados pelo KNNImputer, com o número de vizinhos \(k\) definido via Optuna, seguidos de padronização (\textit{StandardScaler}).
%         \item Variáveis categóricas foram imputadas pela moda (\textit{SimpleImputer}) e codificadas por \textit{One-Hot Encoding}.
%     \end{itemize}
%     \item \textbf{Balanceamento}: Aplicou-se o SMOTE apenas nos dados de treinamento durante a validação cruzada.
%     \item \textbf{Classificação}: Utilização dos algoritmos de machine learning, como XGBoost, LightGBM e CatBoost.
% \end{itemize}

\subsubsection{Metrics}

To evaluate the performance of the classification algorithms, five commonly used metrics in the literature were employed. Accuracy, Balanced Accuracy, Recall, Precision, and F-score are calculated as follows:

\begin{equation}
    ACC = \frac{TP + TN}{TP + FP + TN + FN}
    \label{ACC}
\end{equation}

\begin{equation}
    BACC = \frac{\frac{TP}{TP + FN} + \frac{TN}{TN + TP}}{2}
    \label{BACC}
\end{equation}

\begin{equation}
    Recall = \frac{TP}{TP + FN}
    \label{RE}
\end{equation}

\begin{equation}
    Precision = \frac{TN}{TN + FP}
    \label{PR}
\end{equation}

\begin{equation}
    \textit{F-score} = 2* \frac{Recall * Precision}{Recall + Precision}
    \label{FS}
\end{equation}

% \begin{equation}
%       \text{AUC} = \int_0^1 \text{TPR(FPR)} \, d(\text{FPR})
%       \label{AUC}
% \end{equation}

The variables \textit{TP, TN, FP}, and \textit{FN} correspond to True Positive, True Negative, False Positive, and False Negative, respectively. In the context of medical classifications, it is particularly important that the classifier correctly identifies cancer cases (\textit{TP}), since the patient’s condition may be critical and medical intervention must be immediate.

% As variáveis \textit{TP, TN, FP} e \textit{FN} correspondem a Verdadeiro Positivo, Verdadeiro Negativo, Falso Positivo e Falso Negativo, respectivamente
% % , assim como FPR e TPR correspondem a Taxa de Verdadeiros Positivos e Taxa de Falsos Positivos. 
% No contexto de classificações médicas, é especialmente importante que o classificador identifique corretamente casos de câncer (\textit{TP}), uma vez que a condição do paciente pode ser crítica e a intervenção médica deve ser imediata.

\subsubsection{Hyperparameters setting}
\label{sec-busca-hiperparam}

The hyperparameter search was conducted using Optuna, with the goal of maximizing the mean Balanced Accuracy (BACC) obtained through stratified 5-fold cross-validation. In this procedure, the dataset of each scenario was divided into five subsets (folds) of approximately equal size, while preserving the original class distribution within each fold. In each iteration, four folds were used to train the model and the remaining fold for hyperparameter validation, repeating the process until each fold had served once as the validation set. The mean BACC across the five folds provided an estimate of model performance for the tested hyperparameter configuration. A total of 100 optimization trials per scenario and per model were performed to identify the parameter combination that maximized BACC. The hyperparameter search space is listed in Table~\ref{tbl-hiperparam}.

% A busca de hiperparâmetros foi realizada utilizando o Optuna, com o objetivo de maximizar a média da acurácia balanceada (BACC) obtida por meio de validação cruzada estratificada com 5 \textit{folds}. Nesse procedimento, o conjunto de dados de cada cenário foi dividido em cinco subconjuntos (\textit{folds}) de tamanho aproximadamente igual, mantendo a proporção original de classes em cada fold. Em cada iteração, quatro \textit{folds} foram utilizados para treinar o modelo e o fold restante para validação dos hiperparâmetros utilizados, repetindo-se o processo até que cada \textit{fold} tenha servido uma vez como conjunto de validação. A média da BACC nos cinco \textit{folds} forneceu uma estimativa do desempenho do modelo para a configuração de hiperparâmetros testada. Foram realizadas 100 tentativas de otimização por cenário e por modelo, buscando identificar a combinação de parâmetros que maximizasse a BACC. O espaço de busca de hiperparâmetros está listado na Tabela~\ref{tbl-hiperparam}.

\begin{table}[ht]
    \caption{Hyperparameter search space for XGBoost, CatBoost, LightGBM and KNNImputer.}
    \label{tbl-hiperparam}
    \scriptsize
    \centering
    \begin{tabular}{l|l|l}
        \hline
        Algorithm & Hyperparameter      & Search space               \\ \hline
        \multirow{10}{*}{XGBoost}   & max depth              & {[}3 $\sim$ 15{]}              \\
                                    & number of estimators           & {[}50 $\sim$ 250{]}         \\
                                    & learning rate          & {[}0,01 $\sim$ 0,2{]}       \\
                                    & subsample               & {[}0,5 $\sim$ 1,0{]}         \\
                                    & colsample by tree       & {[}0,6 $\sim$ 1,0{]}         \\
                                    & min child weight      & {[}1,0 $\sim$ 10{]}         \\
                                    & gamma                   & {[}0,0 $\sim$ 1,0{]}        \\ 
                                    & reg alpha              & {[}0,01 $\sim$ 2,0{]}        \\
                                    & reg lambda             & {[}0,01 $\sim$ 5,0{]}        \\
                                    & scale pos weight      & {[}0,5 $\sim$ 5,0{]}         \\ \hline
        \multirow{11}{*}{LightGBM}  & max depth              & {[}3 $\sim$ 15{]}              \\
                                    & number of leaves             & {[}20 $\sim$ 50{]}           \\
                                    & number estimators           & {[}50 $\sim$ 200{]}         \\
                                    & learning rate          & {[}0,01 $\sim$ 0,2{]}       \\
                                    & subsample               & {[}0,5 $\sim$ 1,0{]}         \\
                                    & colsample by tree       & {[}0,6 $\sim$ 1,0{]}         \\
                                    & min child samples     & {[}5 $\sim$ 100{]}        \\
                                    & reg lambda             & {[}0,01 $\sim$ 5,0{]}        \\
                                    & lambda l1              & {[}0,0 $\sim$ 2,0{]}        \\
                                    & lambda l2              & {[}0,0 $\sim$ 5,0{]}        \\
                                    & min split gain        & {[}0,0 $\sim$ 1,0{]}         \\ \hline
        \multirow{7}{*}{CatBoost}   & depth                   & {[}3 $\sim$ 15{]}              \\
                                    & learning rate          & {[}0,01 $\sim$ 0,2{]}    \\
                                    & iterations              & {[}50 $\sim$ 200{]}        \\
                                    & l2 leaf reg           & {[}1 $\sim$ 10{]}               \\ 
                                    & bagging temperature    & {[}0,1 $\sim$ 5,0{]}               \\ 
                                    & border count           & {[}32 $\sim$ 255{]}               \\ 
                                    & random strength        & {[}0,5 $\sim$ 2,0{]}               \\ \hline
        \multirow{1}{*}{KNNImputer} & number of neighbors            & {[}3 $\sim$ 30{]}            \\ 
    \end{tabular}
\end{table}

% \subsubsection{Detalhes de implementação}
% \label{sec-config-exp}

%%=====================================================================================%%

\section{Results}\label{sec:results}

Six binary classification experiments were conducted, applied under two approaches: using the complete dataset and the dataset after dimensionality reduction. The experiments included comparisons between bladder cancer and different urological and oncological conditions, as well as broader scenarios involving multiple pathologies, including comparisons with the baseline \cite{Tsai2022}. For each binary scenario, a standardized computational pipeline was employed, implemented in Python using the Scikit-learn \cite{scikit-learn}, Imbalanced-learn \cite{imbalancedlearn}, and Optuna \cite{akiba2019optuna} libraries.

% Foram conduzidos seis experimentos de classificação binária, aplicados em duas abordagens: utilizando a base de dados completa e a base de dados após redução de dimensionalidade. Os experimentos contemplaram comparações entre câncer de bexiga e diferentes condições urológicas e oncológicas, bem como cenários abrangentes envolvendo múltiplas patologias, incluindo comparações com o baseline \cite{Tsai2022}. Para cada cenário binário, empregou-se um \textit{pipeline} computacional padronizado, implementado em Python com as bibliotecas \textit{Scikit-learn}, \textit{Imbalanced-learn} e Optuna.

The preprocessing stage included the treatment of missing data, where a threshold of 45\% was defined for the removal of variables with high rates of missingness. Variables exceeding this threshold were excluded from the analysis, as in the case of the attribute Calcium in the BC vs. UC experiment, which presented 46.4\% missing values. After this cleaning step, the variable with the highest percentage of remaining missing data in the dataset had 44.2\%. Following data cleaning, the dataset was partitioned into training (80\%) and testing (20\%). To ensure class representativeness in both partitions, stratified sampling was applied.

% A etapa de pré-processamento incluiu um tratamento dos dados ausentes, onde foi definida uma limiar de $45\%$ para remoção de variáveis com alta taxa de ausência. Variáveis que ultrapassaram esse limite foram excluídas da análise, como foi o caso do atributo Calcium no experimento BC vs. UC, que apresentava 46,4\% de valores faltantes. Após essa etapa de limpeza, a variável com a maior porcentagem de dados ausentes remanescente no conjunto de dados possuía 44,2\%. Após a limpeza, o conjunto de dados foi particionado em treinamento (80\%) e teste (20\%). Para garantir a representatividade das classes em ambas as partições, foi empregada a técnica de amostragem estratificada.

\begin{figure}[H]
    \centering
    \subfloat[Bladder Cancer vs. Prostate Cancer]{\includegraphics[scale=0.20]{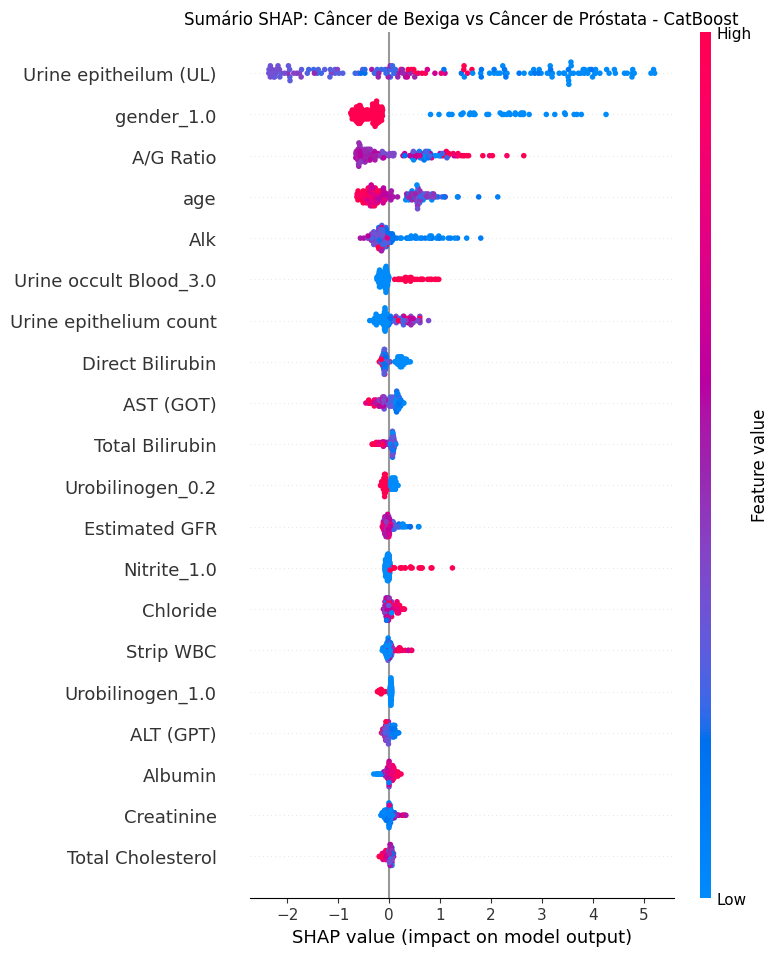}\label{fig:summ_bladder_vs_prostate}}
    \hfill
    \subfloat[Bladder Cancer vs. Cystitis]{\includegraphics[scale=0.20]{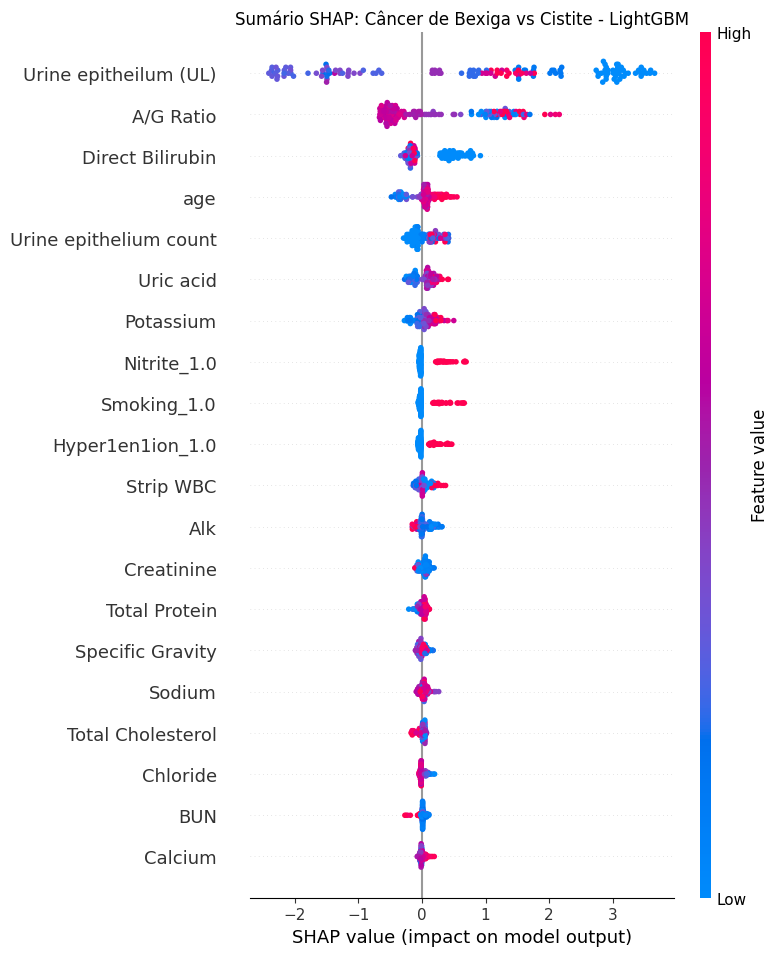}\label{fig:summ_bladder_vs_cistytis}} 
    \hfill
    \subfloat[Bladder Cancer vs. Kidney Cancer]{\includegraphics[scale=0.20]{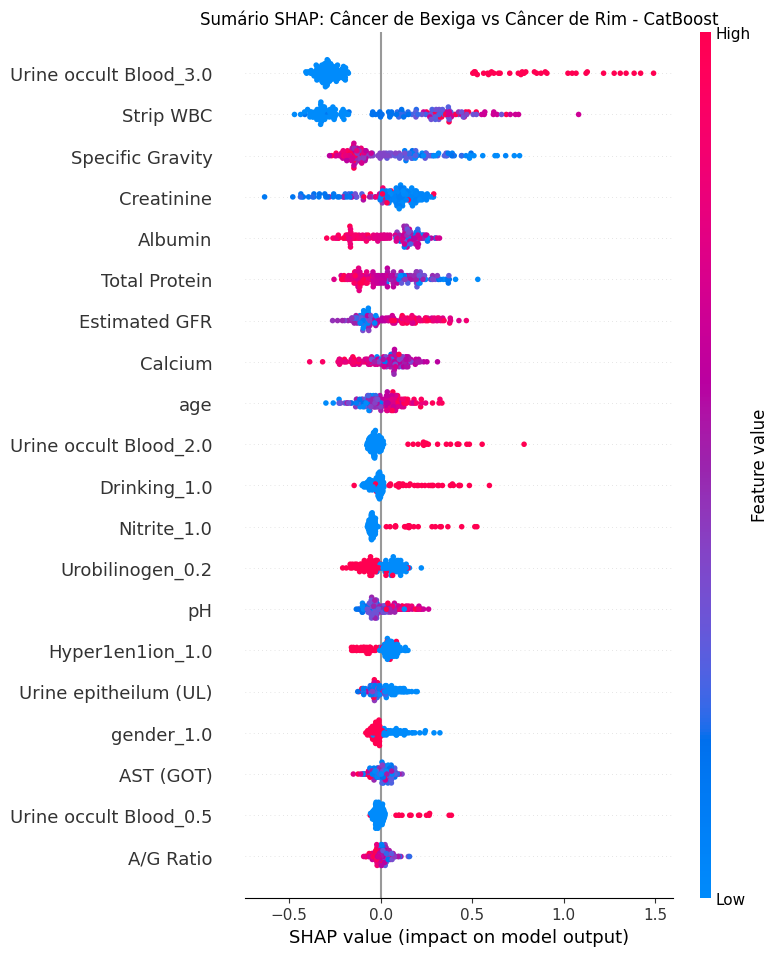}\label{fig:summ_bladder_vs_kidney}} 
    \hfill
    \subfloat[Bladder Cancer vs. Uterus Cancer]{\includegraphics[scale=0.20]{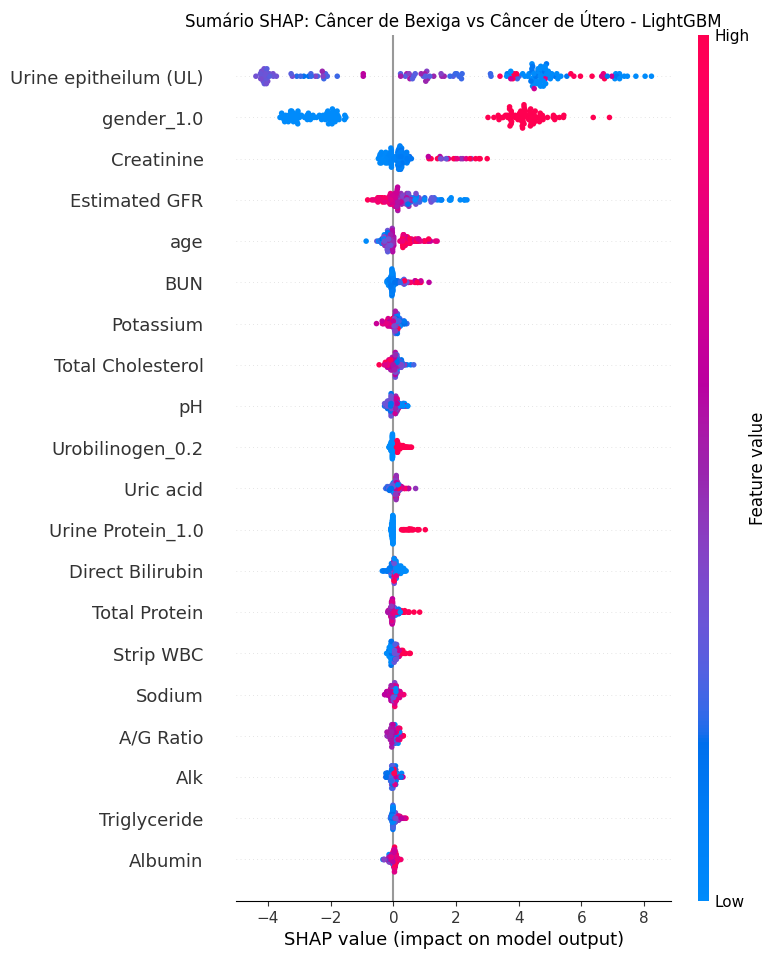}\label{fig:summ_bladder_vs_uterus}} 
    \hfill
    \subfloat[Bladder Cancer vs. Others]{\includegraphics[scale=0.20]{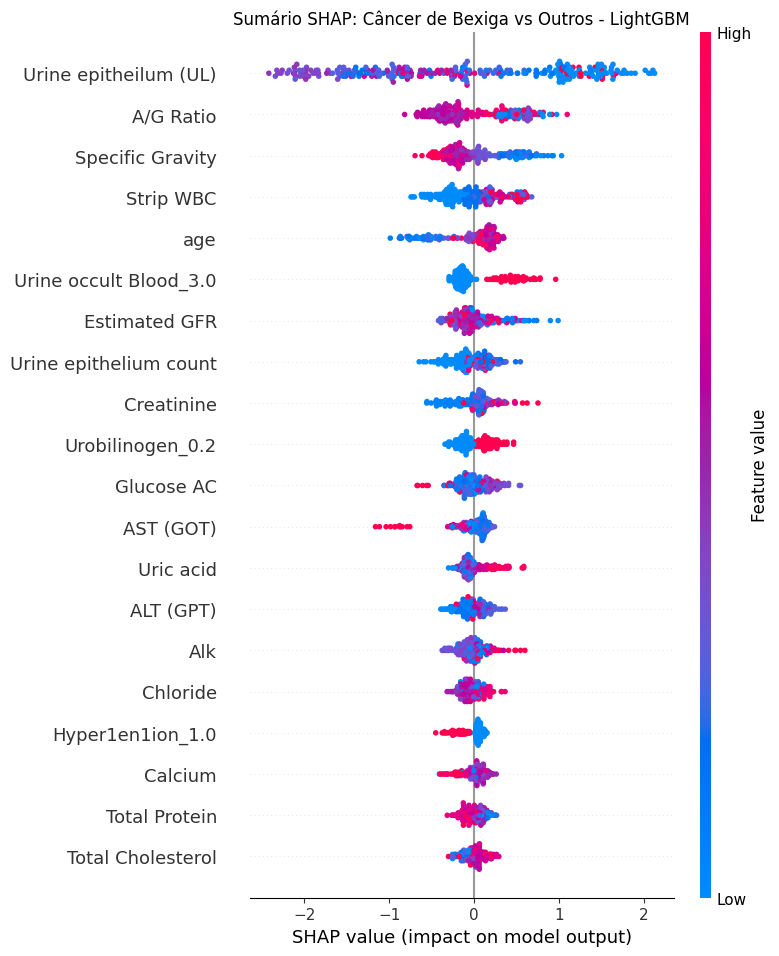}\label{fig:summ_bladder_vs_all}} 
    \hfill
    \subfloat[Prostate Cancer vs. Others]{\includegraphics[scale=0.20]{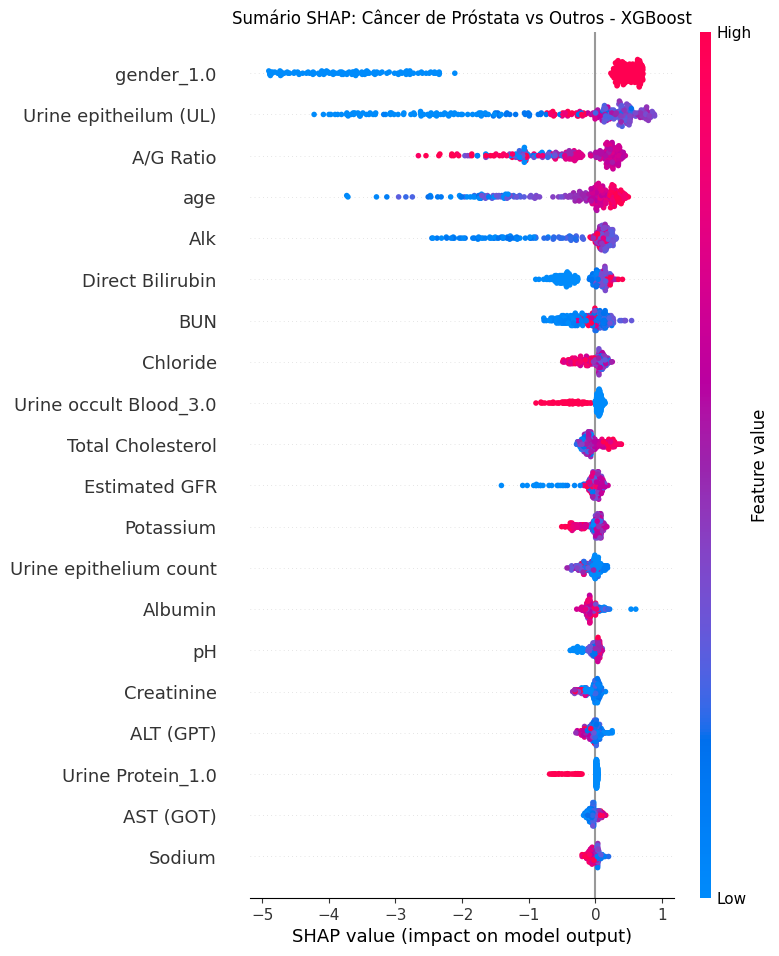}\label{fig:summ_prostate_vs_all}} 
    \caption{SHAP values for the top 20 features of the dataset across all scenarios.}
    \label{fig:shap_all}
\end{figure}

For both approaches, three classification pipelines were constructed, corresponding to the evaluated models (XGBoost, LightGBM, and CatBoost). After hyperparameter optimization with Optuna and evaluation on the test set, the model with the highest Balanced Accuracy (BACC) was selected for each experiment. In the reduced-data approach, an additional step was performed: keeping the same train-test split, SHAP was applied to compute feature importance and assess the impact of dimensionality reduction. The number $N$ of variables with the highest SHAP values was defined as the one that yielded the best BACC, determined through a sensitivity test ranging from 2 up to the total number of available variables in each experiment. The results obtained, with and without feature reduction, are presented in Table~\ref{tbl:classif_all}, while the 20 variables with the highest SHAP values are illustrated in Figure~\ref{fig:shap_all}.

% Para ambas as abordagens, foram construídos três \textit{pipelines} de classificação, correspondentes aos modelos avaliados (XGBoost, LightGBM e CatBoost). Após a otimização de hiperparâmetros via Optuna e a avaliação no conjunto de teste, selecionou-se, em cada experimento, o modelo com maior acurácia balanceada (BACC). Na abordagem com dados reduzidos, realizou-se uma etapa adicional: mantendo a mesma partição treino-teste, aplicou-se o SHAP para calcular a importância das variáveis e avaliar o impacto da redução de dimensionalidade. O número $N$ de variáveis com maior valor SHAP foi definido como aquele que apresentou a melhor BACC, determinado a partir de um teste de sensibilidade variando de 2 até o total de variáveis disponíveis em cada experimento. Os resultados obtidos, com e sem redução de características, estão listados na Tabela~\ref{tbl:classif_all}, enquanto as 20 variáveis com maior valor SHAP são ilustradas na Figura~\ref{fig:shap_all}.

The results indicate that dimensionality reduction had a positive impact in most experiments. In the BC vs. PC experiment, the variables with the highest SHAP values were \textit{Urine epithelium} and \textit{Gender} (Figure~\ref{fig:summ_bladder_vs_prostate}). Since prostate cancer occurs exclusively in male individuals, the model’s attribution of a high SHAP value to this variable is consistent. The best BACC was 95.81\% for the complete model compared to 95.40\% for the model with $N=26$. When compared with the baseline, both the reduced and complete models achieved better accuracy (ACC), with the reduced model yielding the best performance at 95.60\% against 84.80\% from the baseline.

% Os resultados indicam que a redução de dimensionalidade apresentou impacto positivo na maioria dos experimentos. No Experimento CB vs. CP, as variáveis com maior valor SHAP foram \textit{Urine epithelium} e Gênero (Figura~\ref{fig:summ_bladder_vs_prostate}). Como o câncer de próstata ocorre exclusivamente em indivíduos do sexo masculino, observa-se coerência no modelo ao atribuir alto valor SHAP a essa variável. A melhor BACC foi de $95,81\%$ do modelo completo contra $95,40\%$ do modelo com $N=26$ e a comparando com o baseline, a ACC, que o modelo tanto reduzido, quanto completo, obtiveram melhores valores, sendo o melhor o modelo reduzido com $95,60\%$ contra $84,80\%$ do baseline.

In the BC vs. Cystitis experiment, two features stood out: \textit{Urine epithelium} and \textit{A/G Ratio} (Figure~\ref{fig:summ_bladder_vs_cistytis}). A BACC of 97.03\% was obtained with $N=18$, compared to 93.59\% for the complete model. When compared with the baseline, the reduced model achieved 95.24\% against 87.60\%. It is also noteworthy that both precision and specificity reached 100\%.
In the BC vs. KC experiment, the feature \textit{Urine Occult Blood} showed a high SHAP value (Figure~\ref{fig:summ_bladder_vs_kidney}), being a relevant marker in the diagnosis of bladder cancer \cite{Pricee584,dulku2019_hematuria}. The BACC results were 72.02\% obtained with $N=21$, compared to 68.66\% for the complete model. When compared with the baseline, the experiments yielded lower results in terms of ACC and Specificity, with 84.50\% versus 77\% for ACC and 60\% versus 82.9\% for Specificity.

% No Experimento CB vs. Cistite, duas características se destacaram, \textit{Urine epithelium} e \textit{A/G Ratio} (Figura~\ref{fig:summ_bladder_vs_cistytis}). A BACC de $97,03\%$ foi obtida com $N=18$ contra $93.59$ do modelo completo e comparando com o baseline, o modelo reduzido obteve $95.24\%$ contra $87.60\%$ destaca-se também a precisão e especificidade de $100\%$. No Experimento CB vs. CR, a característica \textit{Urine Occult Blood} apresentou alto valor SHAP (Figura~\ref{fig:summ_bladder_vs_kidney}), sendo um marcador relevante no diagnóstico de câncer de bexiga \cite{Pricee584,dulku2019_hematuria}. Os resultados de BACC foram de $72.02\%$ obtida com $N=21$ contra $68.66$ do modelo completo e comparando com o baseline, os experimentos obtiveram resultados inferiores de ACC e Especificidade, sendo $84.50\%$ contra $77\%$ de ACC e $60\%$ contra $82.9$ de Especificidade. 

In the BC vs. UC experiment, \textit{Urine epithelium} once again showed a high SHAP value (Figure~\ref{fig:summ_bladder_vs_uterus}), reflecting the influence of gender in diseases exclusive to females. The two main features were the same as in the BC vs. PC experiment. The best BACC of 96.25\% was obtained with $N=5$, compared to 95\% for the complete model. When compared with the baseline, the reduced model achieved 98.11\% versus 86.90% in ACC.

% No Experimento BC vs. UC, \textit{Urine epithelium} novamente apresentou valor SHAP elevado (Figura~\ref{fig:summ_bladder_vs_uterus}), refletindo a influência de gênero em doenças exclusivas do sexo feminino. As duas principais características foram as mesmas do Experimento CB vs. CP. A melhor BACC de $96,25\%$ foi obtida com $N=5$ contra $95\%$ do modelo completo e comparando com o baseline, o modelo reduzido obteve $98.11\%$ contra $86.90\%$ de ACC. 

In the BC vs. All experiment, the best BACC was 83.46\% for the complete model compared to 83.37\% for the model with $N=43$, once again highlighting the feature \textit{Urine epithelium} Figure~\ref{fig:summ_bladder_vs_all}). In the PC vs. All experiment, the best BACC was obtained with $N=38$, reaching 94.58\% compared to 92.46\% for the complete model, with \textit{Gender} being the most relevant variable according to SHAP values (Figure~\ref{fig:summ_prostate_vs_all}). For these final two experiments, no baseline values were available, and the complete model was used as the reference.

% No Experimento CB vs. All, a melhor BACC foi de $83.46\%$ do modelo completo contra $83.37\%$ do modelo com $N=43$ destacando novamente a característica \textit{Urine epithelium} (Figura~\ref{fig:summ_bladder_vs_all}). No Experimento CB vs. All, a melhor BACC foi obtida com $N=38$, obtendo $94,58\%$ contra $92.46\%$ do modelo completo, sendo a variável de gênero a mais relevante segundo os valores SHAP (Figura~\ref{fig:summ_prostate_vs_all}). Para esses dois experimentos finais, não há valores de baseline disponíveis, sendo o modelo completo utilizado como referência.

% Em diversos casos, como nos Experimentos 2 e 4, a redução resultou em ganhos consideráveis de precisão e especificidade, além de simplificar o modelo.

% \begin{table}[htbp]
\begin{sidewaystable}[htbp]
\caption{Performance achieved by the models in each experiment, considering both the results with the complete dataset and after dimensionality reduction. The Baseline rows refer to the reference values from \cite{Tsai2022}, while the Entire rows indicate the results obtained with the full dataset.}
\centering
\scriptsize
\label{tbl:classif_all}
\begin{tabular}{ccccccccccc}
\hline
\textbf{Exp.} & \textbf{Model} & \textbf{Alg.} & \textbf{\textit{N}} & \textbf{ACC(\%)} & \textbf{BACC(\%)} & \textbf{Prec.(\%)} & \textbf{Sens.(\%)} & \textbf{Spec.(\%)} & \textbf{F1(\%)} \\ \hline
\multirow{3}{*}{BC vs. PC} & Reduced & CatBoost   & 26 & \textbf{95,60} & 95,40 & 98,28 & \textbf{95,80} & 95,00 & \textbf{97,02} \\
 & Entire & CatBoost   & 57  & 94,97 & \textbf{95,81} & \textbf{99,12} & 94,12 & \textbf{97,50} & 96,55 \\
 & Baseline \cite{Tsai2022} & LightGBM   & -  & 84,80 & -     & 86,60 & 84,40 & 85,10 & 85,10 \\ 
\hline
\multirow{3}{*}{BC vs. Cystitis} & Reduced & LightGBM   & 18 & \textbf{95,24} & \textbf{97,03} & \textbf{100,00} & \textbf{94,07} & \textbf{100,00} & \textbf{96,94} \\
 & Entire & LightGBM   & 57  & 93,88 & 93,59 & 98,23 & 94,07 & 93,10 & 96,10 \\
 & Baseline \cite{Tsai2022} & LightGBM   & -  & 87,60 & -     & 86,30 & 89,50 & 85,50 & 87,70 \\ 
\hline
\multirow{3}{*}{BC vs. KC} & Reduced & CatBoost   & 21 & 77,00 & \textbf{72,02} & \textbf{86,21} & \textbf{84,03} & 60,00 & \textbf{85,11} \\
 & Entire & CatBoost   & 57  & 72,96 & 68,66 & 85,19 & 77,31 & 60,00 & 81,06 \\
 & Baseline \cite{Tsai2022} & LightGBM   & -  & \textbf{84,50} & -     & 83,00 & 86,80 & \textbf{82,90} & 84,50 \\ 
\hline
\multirow{3}{*}{BC vs. UC} & Reduced & LightGBM   & 5  & \textbf{98,11} & \textbf{96,25} & \textbf{97,54} & \textbf{100,00} & \textbf{92,50} & \textbf{98,76} \\
& Entire & LightGBM   & 57  & 97,48 & 95,00 & 96,75 & 100,00 & 90,00 & 98,35 \\
& Baseline \cite{Tsai2022} & LightGBM   & -  & 86,90 & -     & 87,10 & 87,80 & 86,70 & 87,30 \\ 
\hline
\multirow{3}{*}{BC vs. All} & Reduced & LightGBM   & 43 & \textbf{83,58} & 83,37 & \textbf{81,51} & 81,51 & \textbf{85,23} & 81,51 \\
 & Entire & LightGBM   & 57  & 83,58 & \textbf{83,46} & 80,99 & \textbf{82,35} & 84,56 & \textbf{81,67} \\
 & Baseline \cite{Tsai2022} & -   & -  & -     & -     & -     & -     & -     & -     \\ 
\hline
\multirow{3}{*}{PC vs. All} & Reduced & XGBoost   & 38 & \textbf{92,54} & \textbf{94,58} & \textbf{67,24} & \textbf{97,50} & \textbf{91,67} & \textbf{79,59} \\
 & Entire & XGBoost   & 57  & 90,67 & 92,46 & 62,30 & 95,00 & 89,91 & 75,25 \\
 & Baseline \cite{Tsai2022} & -   & -  & -     & -     & -     & -     & -     & -     \\ \hline
\end{tabular}
\end{sidewaystable}
% \end{table}

%%=====================================================================================%%

% \section{Results}\label{sec:results}

% Discussions should be brief and focused. In some disciplines use of Discussion or `Conclusion' is interchangeable. It is not mandatory to use both. Some journals prefer a section `Results and Discussion' followed by a section `Conclusion'. Please refer to Journal-level guidance for any specific requirements. 

%%=====================================================================================%%

\section{Discussion}\label{sec:Discussion}

In several experiments, the metrics outperformed those reported by Tsai et al. (2022) \cite{Tsai2022}, particularly in terms of Balanced Accuracy (BACC), precision, and specificity. It can be observed that, in most scenarios, dimensionality reduction either maintained or improved the metrics obtained with the complete dataset.

% Em diversos experimentos, as métricas superaram as reportadas por Tsai et al. (2022) \cite{Tsai2022}, especialmente em termos de acurácia balanceada (BACC), precisão e especificidade. Observa-se que, na maioria dos cenários, a redução de dimensionalidade manteve ou melhorou as métricas obtidas com a base completa. 

In some cases, the removal of redundant variables or those with low SHAP importance resulted in improved performance: in the BC vs. Cystitis experiment, the use of only 18 features increased BACC from 93.59\% to 97.03\%, while in the BC vs. KC experiment the best performance was achieved with only 5 variables, maintaining 100\% sensitivity. This effect can be explained by the mitigation of overfitting risk, since simpler models tend to generalize better in datasets with a limited number of samples. However, in more complex scenarios or those with strong feature overlap between classes, such as the BC vs. KC experiment, dimensionality reduction did not provide significant gains.

% Em alguns casos, a remoção de variáveis redundantes ou com baixa importância SHAP resultou em aumento de desempenho: no Experimento CB vs. Cistite, o uso de apenas 18 características aumentou a BACC de $93.59\%$ para $97.03\%$, enquanto no Experimento CB Vs. CR o melhor desempenho foi alcançado com apenas 5 variáveis, mantendo sensibilidade de $100\%$. Esse efeito pode ser explicado pela mitigação do risco de sobreajuste, já que modelos mais simples tendem a generalizar melhor em bases com número limitado de amostras. No entanto, em cenários mais complexos ou com forte sobreposição de características entre classes, como no Experimento CB Vs. CR, a redução de dimensionalidade não trouxe ganhos expressivos.

It is noteworthy that certain variables consistently proved important across the experiments. For example, the feature \textit{Urine Epithelium} (UL) frequently ranked among the top two features with the highest SHAP values in all experiments except the third. The presence of epithelial cells in urinary cytology, although not a consolidated biomarker for bladder cancer, may be associated with pathological alterations in the urinary tract \cite{LOTAN2003109}. For all scenarios involving gender-specific diseases, the feature \textit{Gender} was correctly ranked with a higher SHAP value. In the scenario involving bladder cancer and kidney cancer, the feature \textit{Urine Occult Blood} obtained the highest SHAP value, which is consistent with the most frequent symptom of bladder cancer, hematuria (blood in the urine) \cite{HalasehSattamA2022}. In this dataset, the manifestation was microscopic hematuria.

% Nota-se que certas variáveis se mostraram comumente importante através dos experimentos, como, por exemplo, a característica \textit{Urine Epitheilum (UL)} se posicionando frequentemente entre uma das 2 características de maior valor SHAP, em todos os Experimentos, exceto o terceiro. A presença de células epiteliais, presente na citologia urinária, embora não seja um biomarcador consolidado para câncer de bexiga, sua identificação pode estar relacionada a alterações patológicas no trato urinário \cite{LOTAN2003109}. Para todos os cenários com doenças exclusivas ao gênero, a característica \textit{gender} foi elencada corretamente com um valor SHAP superior. No cenário contendo câncer de bexiga e câncer de rim, a característica \textit{Urine occult Blood} obteve o maior valor SHAP, estando de acordo com o sintoma mais frequente do câncer de bexiga, que é a hematúria (presença de sangue na urina) \cite{HalasehSattamA2022}, que para este conjunto de dados é a hematúria microscópica.

The application of explainable methods such as SHAP made it possible to identify factors with a potential direct relationship to diagnosis, including laboratory and demographic variables that maintained high predictive weight. The presence of epidemiologically plausible relationships reinforces the usefulness of this approach in assisting specialists with the prioritization of examinations and the formulation of diagnostic hypotheses. It is worth noting that the dataset used presents class imbalance, which, despite the application of SMOTE, may introduce biases. Another point is that although SHAP is effective in assigning importance, it does not explicitly capture dependency relationships, making it necessary to combine it with specific methods.

% A aplicação de métodos explicáveis, como o SHAP, permitiu identificar fatores com possível relação direta com o diagnóstico, como variáveis laboratoriais e demográficas que mantiveram alto peso preditivo. A presença de relações epidemiologicamente plausíveis reforça a utilidade da abordagem para auxiliar especialistas na priorização de exames e na formulação de hipóteses diagnósticas. Vale ressaltar que o conjunto de dados utilizado apresenta desbalanceamento entre classes, o que, apesar da aplicação do SMOTE, pode introduzir vieses. Outro ponto é que, embora o SHAP seja eficaz na atribuição de importância, ele não captura relações de dependência explicitamente, sendo necessário combiná-lo a métodos específicos.

\section{Conclusion}\label{sec:conclusion}

This study proposed a methodology for feature selection to support the diagnosis of bladder cancer, integrating explainability through SHAP (SHapley Additive exPlanations) within a pipeline that combined preprocessing, data imputation, class balancing with SMOTE, and hyperparameter optimization via Optuna, using three machine learning algorithms (XGBoost, LightGBM, and CatBoost). Six binary classification scenarios were conducted, comparing bladder cancer with different urological and oncological conditions as well as broader pathological contexts. Feature selection based on SHAP values effectively reduced dimensionality without significant loss of performance and, in some cases, improved metrics such as balanced accuracy, precision, and specificity. Despite limitations related to limited data availability in certain scenarios and heterogeneity in clinical attribute completion, the results demonstrate that combining explainable methods with robust pipelines is a promising strategy for clinical decision support systems, enhancing both model transparency and interpretability.

% As future work, it is proposed to expand the dataset to include a larger number of patients and clinical conditions, as well as to investigate the applicability of the methodology to other types of cancer and with the addition of samples from patients with negative diagnoses.

% Como trabalhos futuros, propõe-se a ampliação da base de dados para incluir um maior número de pacientes e condições clínicas, assim como a investigação a aplicabilidade da metodologia em outros tipos de câncer e com a adição de amostras de pacientes com diagnóstico negativo.

\backmatter

% \bmhead{Supplementary information}

% If your article has accompanying supplementary file/s please state so here. 

% Authors reporting data from electrophoretic gels and blots should supply the full unprocessed scans for key as part of their Supplementary information. This may be requested by the editorial team/s if it is missing.

% Please refer to Journal-level guidance for any specific requirements.

% \bmhead{Acknowledgements}

\section*{Acknowledgements}

R.A. Krohling thanks the Brazilian research agency Conselho Nacional de Desenvolvimento Científico e Tecnológico (CNPq), Brazil - grant no. 302021/2025-6.

% Please refer to Journal-level guidance for any specific requirements.

\section*{Declarations}

\textbf{Funding} The funders had no role in study design, data collection and analysis, decision to publish, or preparation of the manuscript. \\

\noindent\textbf{Competing interests} The authors declare that they have no known competing financial interests or personal relationships that could have appeared to influence the work reported in this paper. \\

\noindent\textbf{Author contribution} All authors have made substantial contributions to the conception and design, revising the manuscript, and the final approval of the version to be published. Also, all authors agreed to be accountable for all aspects of the work in ensuring that questions related to the accuracy or integrity of any part of the work are appropriately investigated and resolved.

\end{document}